\documentclass[a4paper,10pt,twocolumn]{article}
\usepackage[a4paper,right=1.5cm,left=1.5cm,top=2cm,bottom=2.5cm,headsep=0cm,footskip=0.5cm]{geometry}
\usepackage[spanish,activeacute,es-tabla]{babel}

\usepackage[utf8]{inputenc}
\usepackage{enumitem}
\usepackage{amsmath}
\usepackage{times}
\usepackage{graphicx}
\usepackage{float}
\usepackage{csquotes} 
\usepackage{multirow}
\usepackage{xcolor}
\usepackage{colortbl}
\usepackage[colorlinks=true,allcolors=blue]{hyperref}
\usepackage{titlesec}

\titleformat{\section}{\large\bfseries}{\thesection}{1em}{} % Cambia el tamaño de la fuente de las secciones a \large
\titleformat{\subsection}{\small\bfseries}{\thesubsection}{1em}{} % Cambia el tamaño de la fuente de las subsecciones a \large
\titleformat{\subsubsection}{\small\bfseries}{\thesubsubsection}{1em}{} % Cambia el tamaño de la fuente de las subsubsecciones a \large
\titlespacing{\section}{0pt}{\parskip}{\parskip}
\titlespacing{\subsection}{0pt}{\parskip}{\parskip}

\renewcommand\thesection{\arabic{section}.}
\renewcommand\thesubsection{\arabic{section}.\arabic{subsection}.}

\title{\LARGE \textbf{Initialization matters in few-shot adaptation of vision-language models for histopathological image classification}}

 \author{\large Pablo Meseguer $^{1,2}$, Rocío del Amor $^{1}$, Valery Naranjo $^{1,2}$
 \\
 \\
 \small $^1$ Instituto Universitario de Investigación en Tecnología Centrada en el Ser Humano (HUMAN-Tech), \\ \small Universitat Politécnica de Valencia, Valencia, Spain \{pabmees, madeam2, vnaranjo\}@upv.es \\
 \small $^2$ Valencian Graduate School and Research Network for Artificial Intelligence (valgrAI), Valencia, Spain \\
 }

\date{} % Elimina la fecha

\begin{document}	

\maketitle
\thispagestyle{empty}

\begin{center}
\large
\textbf{Abstract}
\end{center}
\small
\textit{Vision language models (VLM) pre-trained on datasets of histopathological image-caption pairs enabled zero-shot slide-level classification. The ability of VLM image encoders to extract discriminative features also opens the door for supervised fine-tuning for whole-slide image (WSI) classification, ideally using few labeled samples. Slide-level prediction frameworks require the incorporation of multiple instance learning (MIL) due to the gigapixel size of the WSI. Following patch-level feature extraction and aggregation, MIL frameworks rely on linear classifiers trained on top of the slide-level aggregated features. Classifier weight initialization has a large influence on Linear Probing performance in efficient transfer learning (ETL) approaches based on few-shot learning. In this work, we propose Zero-Shot Multiple-Instance Learning (ZS-MIL) to address the limitations of random classifier initialization that underperform zero-shot prediction in MIL problems. ZS-MIL uses the class-level embeddings of the VLM text encoder as the classification layer's starting point to compute each sample's bag-level probabilities. Through multiple experiments, we demonstrate the robustness of ZS-MIL compared to well-known weight initialization techniques both in terms of performance and variability in an ETL few-shot scenario for subtyping prediction.}

\normalsize

% \href{mailto:contacto@caseib.es}{contacto@caseib.es}.

% --> Items
% \begin{itemize}[topsep=0pt, partopsep=0pt, parsep=0pt]
% \item 
% \end{itemize}

\section{Introduction}
The emergence of vision-language supervision \cite{radford2021learning} has shifted the paradigm in many computer vision tasks. This novel form of supervision has permitted the pre-training of vision-language models (VLM) on massive datasets of image-caption pairs. Already pre-trained VLMs allow both zero-shot transfer and transfer learning in downstream tasks thanks to the robust and discriminative representations learned from the data. In particular, research on VLM adaptation has focused on efficient-transfer learning (ETL) where only an additional small set of trainable parameters is updated while the model keeps frozen during training \cite{jia2022visual} and only a few labeled examples are available at adaptation time. The most straightforward technique for ETL is Linear Probing (LP) \cite{radford2021learning}, which involves training a multiclass logistic regression for classification using the features extracted by the VLM image encoder. Recent work on VLM adaptation for natural images \cite{silva2024closer} has proved the limited performance of LP for ETL approaches in few-shot settings, usually underperforming zero-shot (ZS) performance.
\\

In computational pathology (CPath), whole-slide images (WSI) are obtained by digitizing biopsy samples that pathologists use to analyze under the microscope for cancer diagnosis. The gigapixel size of WSIs imposes hardware limitations, restricting their use in computer vision tasks. Weakly supervised learning based on Multiple Instance Learning (MIL) \cite{ilse2018attention} has been widely applied to address this challenge. This approach allows for slide-level classification using only WSI labels, avoiding the need for time-consuming patch-level annotations. Recently, several works have developed VLM tailored for CPath tasks using extensive datasets containing images and textual descriptions from the histopathological domain \cite{huang2023visual,lu2024visual}. MI-Zero \cite{lu2023visual} proposed to adapt the zero-shot transfer ability for slide-level prediction, assessing the class similarities at the patch level. Additionally, task-specific supervised slide-level classification in MIL paradigms has been explored as a downstream \cite{lu2024visual} for subtyping prediction. In particular, an embedding-based MIL framework consists of (1) patch-level feature extraction, (2) feature aggregation at the slide level, and (3) a linear classification layer with softmax activation to compute the slide-level probabilities, which is equivalent to LP. 

As mentioned above, LP has shown performance degradation in ETL few-shot scenarios. Moreover, prior research \cite{lu2024visual} has yet to explore the impact of classifier weight initialization for weakly-supervised MIL tasks. Motivated by these two facts, we propose Zero-Shot Multiple-Instance Learning (ZS-MIL), a simple yet effective method for MIL frameworks that extracts patch-level features using a VLM image encoder and initializes the weights in the classification layer with the zero-shot prototypes. ZS-MIL aims to avoid performance variability associated to random initialization by providing prior knowledge encoded by the text model in the text embeddings of each class. 

\section{Methodology}
In this section, we aim to provide the full details about the proposed approach Zero-Shot Multiple Instance Learning (ZS-MIL). The framework of ZS-MIL is presented in Figure 1. In the following, we include the problem formulation and the description of each component of the framework. \\

\begin{figure*}[hbt]
	\centering
	\includegraphics[width=1.5\columnwidth]{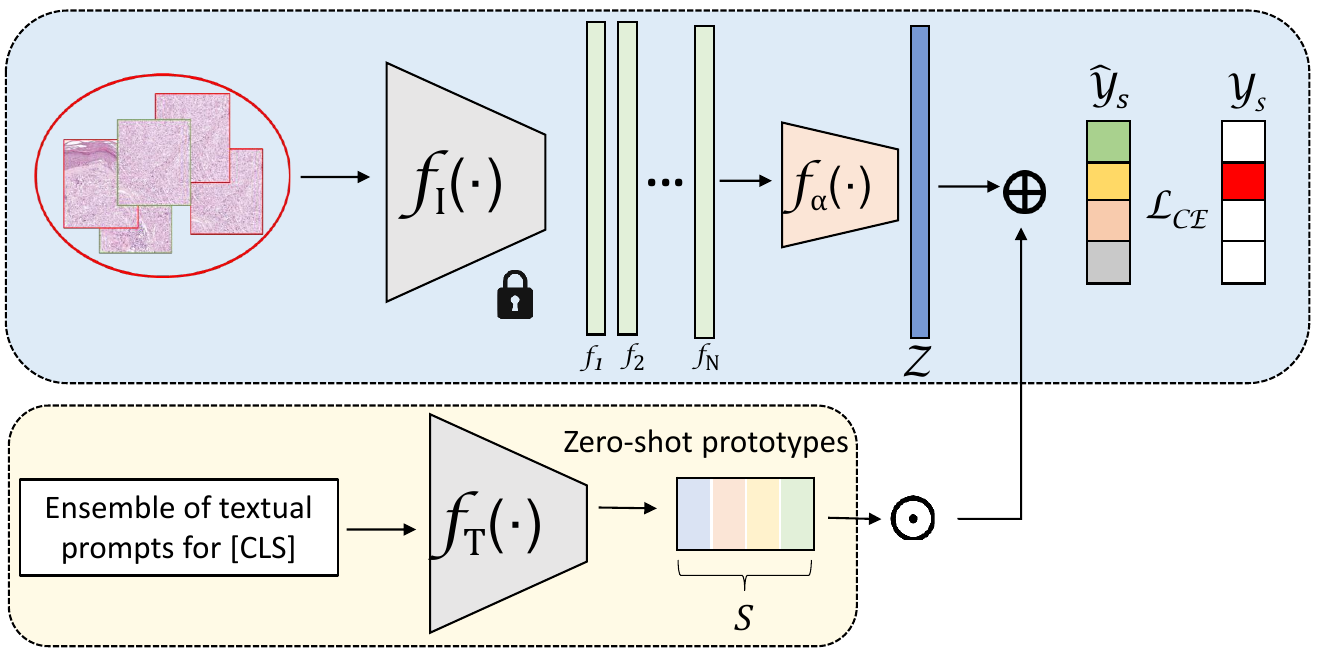}
    \\
	\textit{\textbf{Figure 1.} Overview of Zero-Shot Multiple-Instance Learning (ZS-MIL). ZS-MIL uses an image encoder ($f_{I}$) to extract patch features ($f_N$) and an aggregation model ($f_{\alpha}$) to obtain the bag embedding ($Z$). Textual prompts for the $S$ classes are encoded using the text model ($f_T$) to obtain the zero-shot prototypes for weight initialization on the classification layer.}
	\label{fig_zsmil}
\end{figure*}

\textbf{Problem formulation:} In MIL paradigms, WSIs are arranged in bags ($X$) containing an arbitrary number of instances $N$, such as $X = \{x_n\}_{n=1}^{N}$. Each bag is assigned to one of $S$ mutually exclusive classes ($Y_{s}$). We employ the image encoder of a VLM ($f_{I}$) as patch-level feature extractor, which projects instances to a lower-dimensional manifold $\mathcal{F}$. Subsequently, we define a MIL aggregation function ($f_{\alpha}$), which combines the instance-level projections into a global embedding $Z$. 

\subsection{Zero-Shot Multiple-Instance Learning (ZS-MIL)}
\textbf{Generation of zero-shot prototypes}: As we rely on patch-level feature extraction using the image encoder of a VLM, ZS-MIL aims to exploit the multimodal alignment in the latent space of both image and text encoders by using the zero-shot prior knowledge. For that purpose, we design an ensemble of textual prompts $T$ that describe each of the $S$ classes in the subtyping task to obtain the text embeddings ($w_T$) with the VLM text encoder ($f_T$) as follows:

\begin{equation}
\label{eq_wt}
w_T = f_T(\text{Tokenizer}([T(S_i)]))
\end{equation}

\textbf{Classifier initialization}: In the last stage of conventional MIL paradigms, a randomly initialized multilayer perceptron followed by a softmax activation function is employed to obtain the bag-level class scores. In a few-shot learning scenarios, it exists a high dependence on weight initialization that might hamper performance and underperform zero-shot transfer \cite{silva2024closer}. For that purpose, ZS-MIL proposes to compute the predicted probabilities at the slide level ($\hat{Y}_{s}$) using a multiclass regression classifier initialized with the zero-shot prototypes constructed in Equation \ref{eq_wt} on top of the slide-level aggregated embeddings ($Z$) as follows:

\begin{equation}
\label{eq_wt}
\hat{Y}_c = \frac{\exp(\mathbf{Z} \cdot \mathbf{w}_T^\top / \tau)}{\sum_{i=1}^C \exp(\mathbf{Z} \cdot \mathbf{w}_T^\top / \tau)}.
\end{equation}

where $\tau$ is a temperature parameter. 

Note, that the dot product between the bag embedding and the class prototypes corresponds to the cosine similarity as $l_2$ normalization is applied both for the features and the prototypes. Finally, the optimization of the model parameters, including the weights (if any) in the aggregation module and the text prototypes ($w_T$) is guided by minimizing the standard categorical cross-entropy loss between the reference labels and predicted scores:

\begin{equation}
\label{eq_ce}
\centering
\mathcal L_{CE} = -\frac{1}{S}\sum_{s=1}^{S} Y_{S}log(\hat{Y}_{S})
\end{equation}

where $Y_{s}$ and $\hat{Y}_{s}$ are the bag-level labels and predicted scores, respectively. 

\section{Experiments and results}
\subsection{Dataset}
To validate the proposed approach, we refer to a public dataset from The Cancer Genome Atlats (TCGA) including slides from patients diagnosed with Non-Small Cell Carcinoma (NSCLC). In particular, the dataset contains 445 WSI of lung squamous cell carcinoma (LUSC) and 291 WSI of lung adenocarcinoma (LUAD). We used CLAM \cite{lu2021data} to preprocess the WSIs including tissue segmentation and tile extraction of 256x256 patches at equivalent 20x magnification without overlap.

\subsection{Experimental settings}
The dataset was split into training and testing sets with a 70/30\% ratio. Additionally, 20\% of the development set was used to validate the ablation experiment, while the remaining constituted the pool to select examples for few-shot learning. In particular, we select $k = \{4, 16\}$ as the number of shots per class to evaluate ZS-MIL in high and low-shot scenarios. All experiments were assessed using balanced accuracy and repeated five times to account for variability in sample selection. %Models are trained for 20 epochs with AdamW optimizer and a learning rate of $10^{-4}$.

We additionally include the results on the zero-shot transfer for whole-slide prediction following MI-Zero \cite{lu2023visual}, which computes the average through all patch-level similarities. Note that the zero-shot prototypes obtained following a prompt ensemble approach are the same weights used for classifier initialization in our proposed ZS-MIL.

\subsection{Ablation experiment}
Random initialization of linear classifiers has exhibited performance degradation in linear probing settings \cite{silva2024closer}. This phenomenon is more prominent in few-shot settings due to overfitting the support samples during adaptation. Here, we aim to compare the proposed ZS-MIL with two well-known techniques for weight initialization such as Kaiming \cite{he2015delving} and Xavier \cite{glorot2010understanding}, both in their normal and uniform versions. In the current ablation experiment, we used attention-based aggregation (ABMIL) \cite{ilse2018attention} to obtain the slide-level embedding.

Results shown in Table 1 demonstrate the ability of ZS-MIL to surpass all other methods based on random initialization, improving the second best performing method (Xavier Uniform initialization) by 5.17\% in a high-shot ($k = 16$) and 19.57\% in a low-shot setting ($k = 4$). These results highlight the importance of a carefully initialized classifier, even more, when the number of training samples in the adaptation is limited. Model variability is also addressed with ZS-MIL (standard deviation of 2.44\% and 3.73\%) showing more consistent classification performance despite the selection of different samples for training. 

\begin{center}
\begin{tabular}[hbt]{l c c}
\label{tbl_ablation}
\\ \hline \hline
& k = 4 & k = 16 \\ \hline
Zero-Shot (MI-Zero) \cite{lu2023visual} & \multicolumn{2}{c}{$82.95$} \\ \hline
\textit{Kaiming uniform \cite{he2015delving}} & $60.78$\begin{tiny}$\pm4.24$\end{tiny} & $70.26$\begin{tiny}$\pm2.71$\end{tiny} \\
\textit{Kaiming normal \cite{he2015delving}} & $33.01$\begin{tiny}$\pm5.57$\end{tiny} & $45.83$\begin{tiny}$\pm3.54$\end{tiny} \\
\textit{Xavier uniform \cite{glorot2010understanding}} & $65.79$\begin{tiny}$\pm8.89$\end{tiny} & $82.35$\begin{tiny}$\pm2.63$\end{tiny} \\
\textit{Xavier normal \cite{glorot2010understanding}} & $36.37$\begin{tiny}$\pm7.35$\end{tiny} & $56.60$\begin{tiny}$\pm8.65$\end{tiny} \\
\hline
\textit{ZS-MIL (ours)} & \boldmath{$85.36$\begin{tiny}$\pm2.44$\end{tiny}} & \boldmath{$87.52$\begin{tiny}$\pm3.73$\end{tiny}} \\
\hline
\hline
\end{tabular}
\end{center}
\begin{center}
	\textit{\textbf{Table 1.} Comparison of random weight initialization techniques with ZS-MIL. Results denote the average and standard deviation through five runs.}
\end{center}

\subsection{MIL Efficient Transfer Learning}
Adaptation of VLM focuses on ETL approaches where only an additional small set of parameters is updated in the downstream adaptation stage. As VLMs incorporate large image encoders based on Vision Transformers (ViT) \cite{dosovitskiy2020image}, patch-level feature extraction remains frozen during adaptation to MIL frameworks and aggregation models work on top of tile feature representation. For obtaining the aggregated embedding of the slide, straightforward transformations such as batch global average (BGAP) or max (BGMP) pooling can be utilized. Additionally, trainable modules based on gated-attention (ABMIL) \cite{ilse2018attention} or self-attention transformers (TransMIL) \cite{shao2021transmil} aim to detect the most relevant regions within the WSI.  

In Table 2, we show the performance on the test subset of the proposed ZS-MIL approach for different embedding aggregators. We denote each aggregation method combined with our proposed initialization approach (\textit{ZS-}). Although both non-trainable aggregation methods perform closely, ZS-ABMIL surpasses these aggregation methods by a margin of around 1 and 3\%. It is also worth mentioning the performance degradation of TransMIL especially in low-shot settings ($k = 4$) with a performance drop of 22.22\% compared to ABMIL. These results confirm the tendency of implementing lightweight adaptation strategies in few-shot learning scenarios as ABMIL encompasses 5x fewer parameters (0.39M vs 2.14M) and number of floating operations (22.3 vs 4.0 GFLOPs) than TransMIL. 

\begin{center}
\begin{tabular}[hbt]{l c c}
\label{tbl_ablation}
\\ \hline \hline
& k = 4 & k = 16 \\ \hline
ZS (MI-Zero) \cite{lu2023visual} & \multicolumn{2}{c}{$83.90$} \\ \hline
\textit{ZS-BGMP} & $83.75$\begin{tiny}$\pm2.08$\end{tiny} & $85.53$\begin{tiny}$\pm1.44$\end{tiny} \\
\textit{ZS-BGAP} & $81.26$\begin{tiny}$\pm1.50$\end{tiny} & $83.67$\begin{tiny}$\pm2.05$\end{tiny} \\
\textit{ZS-ABMIL \cite{ilse2018attention}} & \boldmath{$84.16$\begin{tiny}$\pm1.74$\end{tiny}} & \boldmath{$86.48$\begin{tiny}$\pm3.13$\end{tiny}} \\
\textit{ZS-TransMIL \cite{shao2021transmil}} & $61.94$\begin{tiny}$\pm7.54$\end{tiny} & $81.16$\begin{tiny}$\pm2.28$\end{tiny} \\
\hline
\hline
\end{tabular}
\end{center}

\begin{center}
	\textit{\textbf{Table 2.} Efficient-transfer learning results of ZS-MIL for different aggregation models. Results denote the average and standard deviation through five runs.}
\end{center}

\subsection{Qualitative analysis}
Traditionally, artificial intelligent (AI) systems function as black boxes, thus offering no further details into their decision making process. In the medical field, the development of transparent and self-explainable deep learning based models is crucial for their adoption in clinical practice. This approach aims to ensure that AI systems can be trusted by medical professionals by making the decision-making process more transparent and understandable. Explainable artificial intelligence in CPath focuses on the detection of the most critical regions within a tissue sample to provide an explanation for its predictions. Moreover, the selection of region of interest (RoI) promises to guide the pathologist during the evaluation of the tissue sample, thus speeding up and enhancing the diagnostic process. 

Attention-based aggregators such as, Attention-Based MIL \cite{ilse2018attention}, compute an attention score for each patch in a WSI that are used to pool the feature vectors of all instances. These attention scores help in identifying which patches of the tissue sample are most relevant to the diagnosis. In Figure 2, we show the heatmap (Fig. 2B) for a slide with a diagnosis of lung squamous cell carcinoma compared to the pathologist annotation of the tumour regional  (Fig. 2A). The heatmap is colored based on the attention scores with red indicating higher similarity than blue. We can appreciate a notable consistency and overlap between the region delimited by the pathologists and the patches that the trained model has considered for the diagnosis. 

\begin{figure*}[hbt]
	\centering
	\includegraphics[width=1.5\columnwidth]{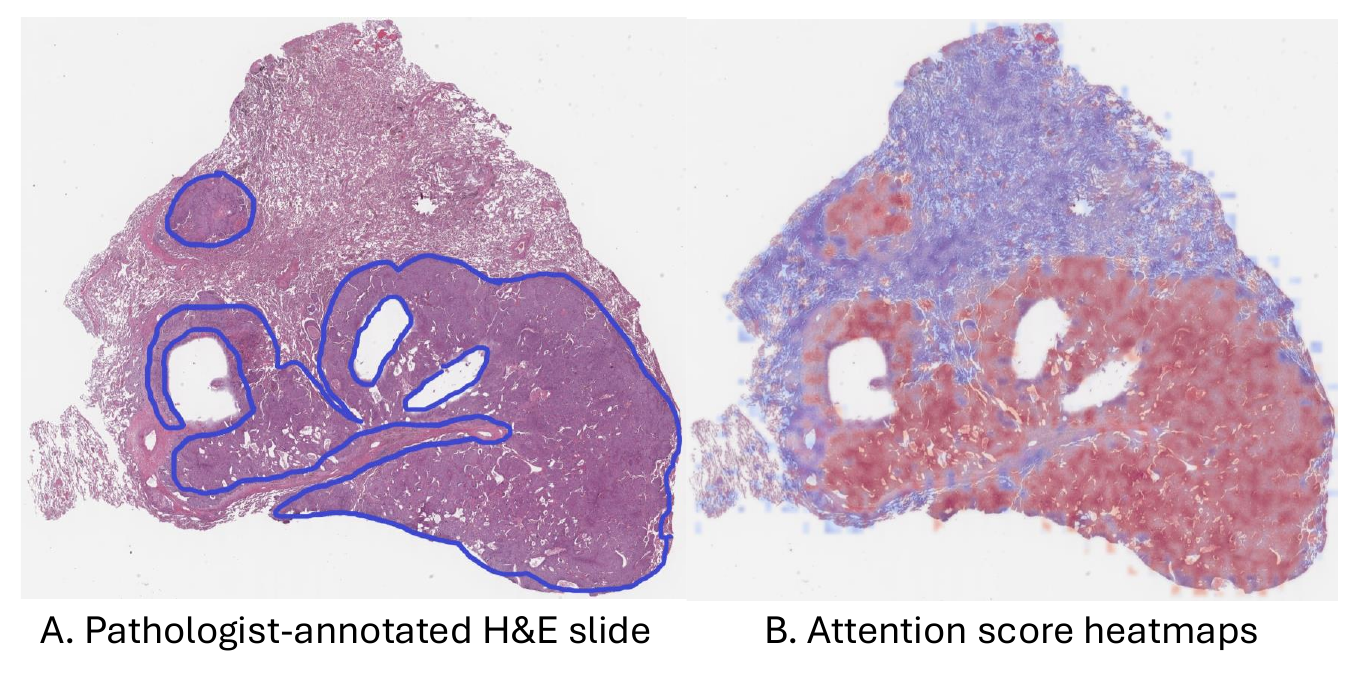}
    \\
	\textit{\textbf{Figure 2.} Intepretability of WSI classification.}
	\label{fig_zsmil}
\end{figure*}

\section{Conclusion}
Adaptation of in-domain vision-language models holds great promise in CPath computer vision tasks due to ability to extract distinctive features from histopathological images. Computational constraints both due to the enormous size of WSI and the large number of parameters makes full model fine-tuning infeasible and encourages work towards ETL approaches. In this work, we demonstrated the performance degradation of few-shot MIL frameworks compared to zero-shot transfer when linear classifiers are randomly initialized despite the implementation of regularization techniques. To address this limitation, the proposed ZS-MIL offers a simple yet effective solution by exploiting the knowledge of the multimodal VLM. ZS-MIL uses the text embeddings of each class in the task at hand to initiate the weights in the final classification layer. We showed that this approach is effective in ETL scenarios through multiple aggregation functions, specially for lightweight models that are less prone to overfitting. Further research lines should work toward the investigation of the explainability of these models and how the inherent knowledge in the encoded text description is conditioning the discovery of regions of interest.

% \small
% \textbf{Funding:} This work has received funding from the Spanish Ministry of Economy and Competitiveness through projects PID2019-105142RB-C21 (AI4SKIN) and PID2022-140189OB-C21 (ASSIST) and from the Generalitat Valenciana through project COM-TACTS2 (CIPROM/2022/20). The work of Rocío del Amor and Pablo Meseguer has been supported by the Spanish Ministry of Universities under an FPU Grant (FPU20/05263) and valgrAI - Valencian Graduate School and Research Network of Artificial Intelligence, respectively. 

% ---- Bibliography ----
\begin{small}
\bibliographystyle{ieeetr}
\bibliography{references}
\end{small}

\end{document}